\documentclass[11pt,a4paper]{article}
\usepackage[hyperref]{emnlp2020}
\usepackage{times}
\usepackage{latexsym}

\usepackage{xspace}   % spacing
\usepackage{multirow}
\usepackage{booktabs}
\usepackage{array}
\usepackage{arydshln}
\usepackage{graphicx}
\usepackage{soul}   % highlighting

\usepackage{xcolor}

\newcommand{\hlc}[2][yellow]{{%
    \colorlet{foo}{#1}%
    \sethlcolor{foo}\hl{#2}}%
}

\usepackage[shortlabels]{enumitem}
\setlist[enumerate]{nosep}

\usepackage{microtype}

\aclfinalcopy % Uncomment this line for the final submission
\def\aclpaperid{***} %  Enter the acl Paper ID here

\newcommand\numtldrs{5,411\xspace}
\newcommand\numpapers{3,229\xspace}

\newcommand{\dataset}{\mbox{\textsc{SciTldr}}\xspace}

\newcommand{\aic}{AIC\xspace}

\newcommand{\ours}{\textsc{Catts}\xspace}
\newcommand{\oursxsum}{\textsc{Catts}$_\mathrm{XSUM}$\xspace}
\newcommand{\bart}{\textsc{BART}\xspace}
\newcommand{\bartxsum}{\textsc{BART}$_\mathrm{XSUM}$\xspace}
\newcommand{\bertsumext}{\textsc{BertSumExt}\xspace}
\newcommand{\fullname}{\underline{C}ontrolled \underline{A}bstraction for \underline{T}LDRs with \underline{T}itle \underline{S}caffolding\xspace}

\newcommand\auth{\tldr-Auth\xspace}
\newcommand\peerreview{\tldr-PR\xspace}
\newcommand\tldr{\textsc{tldr}\xspace}
\newcommand\tldrs{\textsc{tldr}s\xspace}

\title{\textsc{Tldr}: Extreme Summarization of Scientific Documents}

\author{Isabel Cachola$^\dag$ \hspace{1.4em}
Kyle Lo$^\dag$ \hspace{1.4em}
{\bf Arman Cohan$^\dag$} \hspace{1.4em}
{\bf Daniel S. Weld$^{\dag\ddag}$} \vspace{8pt}\\
  $^\dag$Allen Institute for AI  \vspace{2pt} \\
  $^\ddag$Paul G. Allen School of Computer Science \& Engineering, University of Washington \vspace{2pt}\\
  {\small{\tt \{isabelc,kylel,armanc,danw\}@allenai.org}}
}

\date{}

\begin{document}
\maketitle

\begin{abstract}
We introduce \tldr generation, a new form of extreme summarization, for scientific papers.  \tldr generation involves high source compression and requires expert background knowledge and understanding of complex domain-specific language.  To facilitate study on this task, we introduce \dataset, a new multi-target dataset of 5.4K \tldrs over 3.2K papers. \dataset contains both author-written and expert-derived \tldrs, where the latter are collected using a novel annotation protocol that produces high-quality summaries while minimizing annotation burden. We propose \ours, a simple yet effective learning strategy for generating \tldrs that exploits titles as an auxiliary training signal. \ours improves upon strong baselines under both automated metrics and human evaluations. Data and code are publicly available at \href{https://github.com/allenai/scitldr}{https://github.com/allenai/scitldr}.

% \footnote{Code \& data: \url{https://github.com/allenai/scitldr}} 

% KYLE --- I realllly really dislike this.  we have no proof this is that justified, only anecdotal evidence & argument by speculation
% Unlike most summarization datasets, that include a single gold summary for each document, \dataset consists of multiple \tldrs for each paper. This is a critical property for evaluation in high-compression setting where there is natural variation in human-written summaries.
\end{abstract}

\section{Introduction} \label{sec:introduction}
\begin{figure}[t!]
    \centering
   \includegraphics[width=1\linewidth]{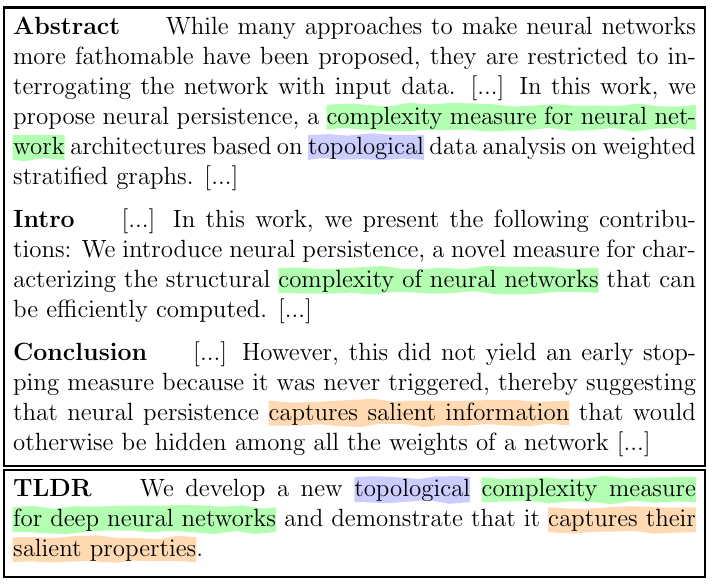}
    \caption{
    An example \tldr of a scientific paper. 
    A \tldr is typically composed of salient information (indicated by colored spans) found in the abstract, intro, and conclusion sections of a paper.} 
    \label{fig:example} 
\end{figure}

% % % % % % % % % % % % % % % % % % 
We introduce \tldr\footnote{\tldr is an acronym that stands for ``too long; didn't read,'' which is often used in online informal discussion (e.g., Twitter or Reddit) about scientific papers. For visual clarity, we omit the semi-colon.} generation for scientific papers. An alternative to abstracts, \tldrs focus on the key aspects of the paper, such as its main contributions, eschewing nonessential background or methodological details.  Given the increasing pace of publication \cite{van2014global} and resulting difficulty in keeping up with the literature, \tldrs can enable readers to quickly discern a paper's key points and decide whether it's worth reading. 
The goal of existing work in summarization of scientific documents is to generate abstracts or provide complimentary summaries to abstracts. \cite{Collins2017ASA,Cohan2018ADA,scisumm-chandrasekan,Yasunaga2019ScisummNetAL}. In contrast, \tldr generation seeks to produce an extreme (single sentence) summary \cite{narayan-cnn-xsum} given the entire paper. Further, \tldr generation is a challenging natural language generation task. Writing a \tldr of a scientific paper requires expert background knowledge and understanding of complex domain-specific language to identify the salient aspects of the paper, while maintaining faithfulness to the source and correctness of the written summary.  An example \tldr is provided in Figure \ref{fig:example}.

% \kyle{\tldr generation is different from past automatic summarization of scientific papers \cite{Collins2017ASA,Cohan2018ADA,scisumm-chandrasekan,Yasunaga2019ScisummNetAL} as existing work mostly focuses on long summaries that are in form of abstracts or complimentory to abstracts. \tldr generation is more similar to extreme summarization \cite{narayan-cnn-xsum} and headline generation \cite{vasilyev2019headline,Tan2017FromNS} [CITE MORE] which also seek to produce single phrase or sentence summaries of entire documents, requiring a lot of compression and abstraction - maybe.  These have only been studied so far on news articles, while \tldr generation pushes this compression stuff further because (i) scientific papers are way longer than news articles, and (ii) focal points are different.} 
% \ac{this is great, but too long. I tried to insert a sentence in the previous paragraph to convey the points. Will move additional info in this paragraph to related work}

To facilitate the study of \tldr generation, we introduce \dataset, a new dataset of \numtldrs \tldrs of computer science papers.
\dataset is built from a combination of \tldrs written by authors of submissions on OpenReview\footnote{\href{https://openreview.net/}{https://openreview.net/}}  and \tldrs derived by a novel annotation protocol that asks domain experts to rewrite peer review comments for that submission.
% \kyle{This gives us multiple gold summaries per paper, which is especially important for [Can we exactly articulate how this is helpful?].}
Having multiple gold summaries per paper is especially important for evaluation when there is variability in human-written gold summaries \cite{zechner1996fast,harman-over-2004-effects}.

In addition to establishing strong extractive and abstractive summarization baselines using Transformer-based \cite{transformers} models, we present \ours (\fullname), a simple yet effective learning strategy for \tldr generation. 
\ours incorporates ideas from scaffold tasks for multitask learning \cite{swayamdipta-etal-2018-syntactic,cohan-etal-2019-structural} and control codes in conditional language generation \cite{Keskar2019CTRLAC} to address the problem of data scarcity in the highly-specialized scientific domain. In particular, \ours exploits titles as an auxiliary, naturally-occurring training signal by training the model to generate both titles and \tldrs indicated by control codes. 
We show that \ours applied to \bart \cite{Lewis2019BARTDS}, a state-of-the-art summarization model, results in performance improvement in both automated metrics and human evaluation. 

% KYLE -- I think what I have above in RED is better.
% We next present a method for generating \tldrs of scientific papers using\Bug{we can't just 'use' BART - need to emphasize additional objective and shuffling} BART \cite{Lewis2019BARTDS}, a pretrained language model with strong performance on summarization. \ac{sentence is weak, could be improved} Drawing upon connections between \tldrs and a related task, title generation, we propose a multitask learning strategy using a title generation scaffold \cite{swayamdipta-etal-2018-syntactic, cohan-etal-2019-structural} \kyle{rethink citation here; never mentions CTRL at all but arguably its more relevant.} for improving \tldr generation by finetuning pretrained language models. We show that this method, while simple, is effective in extreme summarization for the scientific domain. 

% \kyle{What do you think of this bolding of contributions header?  Makes it easier to see, but less flow w/ reading}\isabel{I don't think it's necessary, the numbering already makes it easy to find}
% \paragraph{Our contributions:}
\noindent Our contributions are summarized below: 
\begin{enumerate}[noitemsep,leftmargin=12pt,wide=6pt]
    \item We introduce \tldr generation, a new form of extreme summarization, for scientific papers.  With extensive analysis of properties of \tldrs, we provide insight into the types of information and amount of variability in human-written \tldrs.
    % \kyle{Dan wants to add a contribution point about how we did this analysis to understand \tldrs.  I can see this as separate or attached to either the Task contrib or the Dataset contrib.  }
    \item We release \dataset, a new multi-target dataset of \numtldrs \tldrs over \numpapers scientific papers. \dataset contains both author-written and expert-derived \tldrs, where the latter are collected using a novel annotation protocol that produces high-quality summaries while avoiding the burden of reading the full paper.
    % \kyle{Dan wants us to mention a novel annotation protocol used to circumvent the burden of having to read full papers.}\ac{added a phrase about it, but we may need to improve it} \isabel{edited, i think it's good now?}
    % \Bug{what is number of tldrs, not just papers. also say this in the summary table and elsewhere}
    % \isabel{"OpenReview peer review" sounds odd. Can we omit OpenReview?}
    \item We establish strong baselines on \dataset and improve them with \ours, a simple yet effective learning strategy for generating \tldrs that uses titles as an auxiliary training signal. 
    % \kyle{Is \ours the learning technique or the final model?} \isabel{the technique? we can't claim bart}
    \item We perform extensive analysis and  human evaluation of system-generated \tldrs, focusing on informativeness and factual correctness.
    % \item We propose a novel approach for producing \tldrs, that pretrains on title generation, and investigate several alternative methods. 
    % \item We present experiments showing that 1) intermixing title prediction with \tldr-generation provides a strong boost, and 2) our method outperforms strong baselines. 
\end{enumerate}

\section{Dataset construction}\label{sec:dataset}

% ----- moving around to get onto page 2
\begin{figure}[t]
\centering
\includegraphics[width=1.0\linewidth]{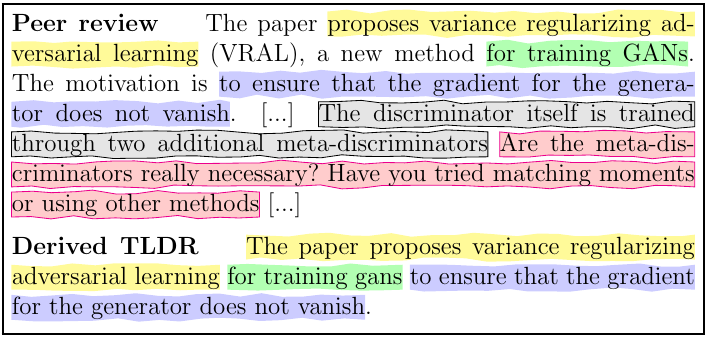}
    \caption{Example of a reviewer comment rewritten as a \tldr  (best viewed in color).  A peer review comment often begins with a summary of the paper which annotators use to compose a \tldr.  Annotators are trained to preserve the original reviewer's wording when possible (indicated by colored spans), and to avoid using any \hlc[gray!10]{\emph{excess details}} or \hlc[red!20]{\emph{criticism}}.}
    \label{fig:review-comment-example} 
\end{figure}
%  -----

    \paragraph{Overview}
    We introduce \dataset, a new multi-target dataset of \numtldrs \tldrs over \numpapers scientific papers in the computer science domain.\footnote{See Appendix Table~\ref{tab:venue_breakdown} for full venue breakdown.}
    % The dataset includes the following breakdown of venues: ICLR (85.2\%),  NeurIPS (5.8\%), OpenReview (2.1\%),  ICML (2.0\%),  other (4.9\%). 
    The training set contains 1,992 papers, each with a single gold \tldr.  The dev and test sets contain 619 and 618 papers each, with 1,452 and 1,967 \tldrs, respectively. This is unlike the majority of existing summarization datasets that assume only one gold summary for a given document.

% ------ moving around to get onto page 3
\begin{table*}[t!]
    \centering
    \small
    \setlength{\tabcolsep}{2pt}
    \renewcommand{\arraystretch}{1.1}
    \begin{tabular}{@{}lrrrrrr@{}}
    \toprule
    \textbf{Dataset}  &
    \begin{tabular}{@{}r@{}}
          \textbf{Number of} \\ \textbf{documents}
    \end{tabular} &
    \begin{tabular}{@{}r@{}}
          \textbf{Avg. words} \\ \textbf{in document}
    \end{tabular} &
    \begin{tabular}{@{}r@{}}
         \textbf{Avg. words} \\ \textbf{in summary}
    \end{tabular}  &
    \begin{tabular}{@{}r@{}}
          \textbf{Compression} \\ \textbf{ratio}
    \end{tabular} & 
    \begin{tabular}{@{}r@{}}
          \textbf{\% novel}\\
          \textbf{words}
    \end{tabular} & 
    \begin{tabular}{@{}r@{}}
          \textbf{Multi-target}
    \end{tabular} \\
    \midrule
    \emph{Non-scientific documents} \\
    \hdashline[0.5pt/0.5pt]\noalign{\vskip 0.5ex}
    DUC \cite{duc2003} & 624 & 441 & 11 & 40.1 & 30.0 & yes \\ 
    NYTimes \cite{sandhaus2008new} & 655K & 549 & 40 & 13.7 & 20.1 & no  \\  % 26.7
    DailyMail \cite{hermann2015teaching} & 220K & 653 & 55 & 11.9 & 17.0 & no \\
    CNN \cite{hermann2015teaching}& 93K & 760 & 46 & 16.5 & 16.8 & no \\
    XSUM \cite{narayan-cnn-xsum}& 226K & 431 & 23 & 18.7 & 35.8 & no \\
    Newsroom \cite{newsroom} & 1.32M & 659 & 27 & 24.4 & 26.0 & no  \\  % 26.7
    BigPatent \cite{Sharma2019BIGPATENTAL} & 1.34M & 3.6K & 117 &  30.5  & 13.6 & no \\
    % [0.5ex]
    \midrule
    \emph{Scientific documents} \\
    \hdashline[0.5pt/0.5pt]\noalign{\vskip 0.5ex}
    CLPubSum \cite{Collins2017ASA} & 10.3K & 8.2K & 226 & 36.5 & 7.7 & no \\ 
    PubMed \cite{Cohan2018ADA} & 133K & 3K &  203 & 14.9 & 10.5 & no \\ 
    ArXiv \cite{Cohan2018ADA} & 215K & 4.9K & 220 & 22.5 & 8.3 & no \\ 
    SciSummNet$^\dagger$ \cite{Yasunaga2019ScisummNetAL} & 1.0K & 4.7K & 150 & 31.2 & 7.4 & no \\   % 31.15
    TalkSumm$^\ddagger$ \cite{Lev2019TalkSummAD} & 1.7K & 4.8K & 965 & 5.0 & 16.5 & no \\ 
    \hdashline[0.5pt/0.5pt]\noalign{\vskip 0.5ex}
    % \dataset (this work) \\
    %  \textbf{\dataset}$_\mathrm{\ao}$  (ours) & 3.2k & 159 & 19 & 8.3 & 43.9 & yes \\
    % \hspace{1em}\aic & 3.2k & 993 & 19 & 52.3 & multi \\
    % \hspace{1em}$\mathrm{\dataset_{Full}}$  & 3.2k & 5,009 & 19 & 263.6 & multi \\
    % \textbf{\dataset}$_\mathrm{\aic}$ (ours) & 3.2K & 993 & 19 & 52.3 & 26.0 & yes \\   
    \textbf{\dataset} (ours)  & 3.2K & 5K & 21 & 238.1 & 15.2 & yes \\
    \bottomrule
    \end{tabular}
    \caption{Comparison of \dataset to existing summarization datasets.  \emph{(i)} \dataset provides multiple summary targets unlike other recent summarization datasets. \emph{(ii)} \dataset requires both extreme compression and abstraction, as evidenced by the compression ratio and novelty (\% of summary words not in the source document), especially when compared with other scientific summarization datasets. 
     \\
    [1mm]
    \footnotesize{$^\dagger$SciScummNet data was later included in the CL-SciSumm shared task and dataset \cite{jaidka2018insights, scisumm-chandrasekan}, which has an additional 40 manually annotated documents and its statistics are similar to SciSummNet.
    } \\
    [1mm]
    \footnotesize{$^\ddagger$Unlike the other summarization datasets presented here, TalkSumm is an automatically-constructed dataset for training; the TalkSumm-supervised model in \citet{Lev2019TalkSummAD} was evaluated using CL-SciSumm \cite{jaidka2018insights}.}
    % $\mathrm{\dataset_{Full}}$ refers to \dataset with full-text while \ao only includes abstract and \aic includes abstract, introduction and conclusion sections.
   } \label{tab:compare-datasets}
\end{table*}

    As evidenced by earlier work in summarization evaluation \cite{Cohan2016RevisitingSE}, variability in human-written summaries \cite{zechner1996fast,harman-over-2004-effects} can negatively impact the reliability of automated summarization metrics like Rouge \cite{lin-2004-rouge}.\footnote{While Rouge is capable of handling multiple targets for a given document, most summarization datasets are single target. See Table~\ref{tab:compare-datasets}.}
    Considering only one gold \tldr for each paper as a basis of automated evaluation might result in inaccurate system quality assessment because content that might appear in a \tldr can have large variability. In addition, having multiple gold summaries for each document enables performing more in-depth analysis and thorough evaluation \cite{nenkova2004evaluating}.
    % \ac{would be nice if we can show this is the case} \kyle{if we cant demonstrate correlation with human judgments; we can reframe value qualitatively and say it can support future research into automated metrics}
    
    To address this, \dataset contains \tldrs written from the perspective of the author (``\auth'') and \tldrs written from the perspective of the peer reviewer(``\peerreview''). We describe these two types of \tldrs in the following paragraphs.
    
    \paragraph{Collecting \auth pairs}
    Scholar-written \tldrs of scientific papers are available on various online platforms.  On OpenReview.org, a publicly available scientific reviewing platform, authors submit \tldrs of their papers that summarize the main content for both reviewers and other interested scholars. Scholars also share \tldrs social media platforms, such as Twitter and Reddit. % For \dataset, we focus on OpenReview and leave collection from social media for future work.\footnote{Initial attempts to use \tldrs on Twitter proved difficult. \tldrs can be spread out across multiple posts, include or reference attached figures, and place substantial emphasis on hashtags and author or affiliation mentions as opposed to content summarization.}\ac{this should go to related work, not here}\isabel{i don't think we should mention that we looked into using twitter data, just say leave it for future work}
    
    We use the OpenReview API\footnote{\href{https://github.com/openreview/openreview-py}{https://github.com/openreview/openreview-py}} to collect pairs of papers and author-written \tldrs, along with the full-text PDFs\footnote{A small fraction of those papers ($<\textrm{5}\%$) did not have an available PDF file, so we could not parse their full body text. This are still included the dataset as it is possible to generate a \tldr from an abstract alone.} of those papers. We use the S2ORC pipeline  \cite{lo-wang-2020-s2orc} to convert PDFs to structured, machine-readable full text.  We then split the papers randomly into the previously-mentioned train, dev, and test sets; each paper at this point has an associated author-written gold \tldr.

    \paragraph{Rewriting peer reviews into \peerreview pairs}
    % \paragraph{Additional \tldrs from peer reviews}
    \label{sec:pr-tldr}
    % Most datasets are single target.
    % Rouge \cite{lin-2004-rouge} is a popular automated evaluation method for summarization used in many datasets. 
    
    %  To address this challenge, we collect additional gold summaries for papers in the dev and test sets. 
    
    % While desirable, s
    Scaling up data collection in a specialized scientific domain is costly and challenging. 
    % Writing faithful \tldrs of scientific papers requires domain expertise, which precludes hiring crowdworkers. Additionally, summarizing scientific papers is time-consuming for domain experts, which makes annotation prohibitively expensive. 
    % To ensure high-quality summaries while minimizing annotation cost and effort, 
    To sidestep this problem, we use a novel annotation protocol that exploits natural summaries in peer review comments.  
    Assuming the typical peer reviewer has carefully scrutinized the source paper and provided a faithful summary in their comment (often in the first paragraph), domain experts can rewrite these comments into \tldrs.
    % (we call these ``\peerreview''). 

% KYLE -- Keeping these as notes, but dont seem important details
% Compared with author-written \tldrs, these \tldrs are from the perspective of a  reader. 
% Having multiple targets allows us to capture the natural variation inherent in summarization and particularly in extreme summarization. 
% We found that the summary statements in the first paragraph of reviews are not easily identifiable by automatic means and further, they are often more verbose than a typical \tldr. 

    For this task, we recruit 28 undergraduate computer science students from the University of Washington with self-reported experience in reading scientific papers.  Each recruited student received one hour of one-on-one writing training and then was asked to work independently. Annotators were only shown the first 128 words of a sampled\footnote{Multiple peer review comments can be available for each paper on OpenReview.  We focused on ensuring that each paper in dev and test had at least one \peerreview.} peer review comment.  They were instructed to keep their \tldrs between 15-25 words (similar to the length of an author written \tldr) and to skip reviews that do not contain a summary or if they did not understand the content. They were also instructed to use the original language in the review, when possible. We manually assessed every written summary, discarding \tldrs that did not adhere to the guidelines, and allowed 20/28 students who performed well to continue work beyond the first hour. Students were compensated at the local median hourly wage of \$20 USD per hour.
    % For this task, we recruit 28 university students majoring in computer science and with self-reported experience reading scientific papers. We provide each with an hour of one-on-one training for the task.  20 of the 28 students who performed well were hired to perform annotations at the local median hourly wage of \$20 USD per hour. 
    % \kyle{Dan, is this OK wrt anonymity.}
    %of \$20 USD per hour.
    %  Washington State
    Refer to Appendix \S\ref{sec:annotation_instructions} for full annotation instructions. Figure~\ref{fig:review-comment-example} contains an example of a peer review and its corresponding \peerreview.  We discuss differences between \peerreview and \auth throughout Section \ref{sec:task}.

    % \arman{what do we do if it is not factually correct? Also we should provide some numbers on annotation quality, agreement, etc. What if the review doesn't have a summary?}. \isabel{we either edit the tldr or throw it out, depending on how much editing it needs}

\section{Dataset analysis}\label{sec:task}

\subsection{Compression and abstractiveness}
Table~\ref{tab:compare-datasets} compares \dataset with other summarization datasets in both scientific and non-scientific domains.
We observe that \dataset has short summaries, like XSUM and NewsRoom, with long source documents, like BigPatent and the other scientific-domain datasets.  This results in a much higher compression ratio compared with existing datasets.  Summarization in higher compression settings is challenging as it requires capturing more precisely the salient aspects of the document \cite{newsroom}. 

Following \citet{narayan-cnn-xsum,newsroom}, we measure abstractiveness (or novelty) by percentage of words in the summary that do not appear in the source document. We observe that \dataset is more abstractive compared with other scientific domain datasets but less abstractive compared with non-scientific domain datasets. We also observe that \dataset is smaller in comparison to automatically collected datasets, such as XSUM and ArXiv, but is larger in comparison to other manually collected datasets, such as SciSummNet.

\subsection{Information content}
\label{sec:information_content}

We analyze the information content of \tldrs using an approach motivated by the nugget-based summarization evaluation framework of \citet{nenkova2004evaluating}.  In a similar manner, we asked two computer science researchers to read through a collection of \tldrs to both define a comprehensive set of categories of types of information present in \tldrs, which we refer to as nuggets.\footnote{While we adopt the term `nugget'' for convenience, we recognize that that they traditionally correspond to factoids, while here they correspond to discourse roles \citet{Teufel1999ArgumentativeZI}.}  We also label each \tldr with all represented nuggets.  Table~\ref{tab: tldr-categories-with-pcts} presents this categorization, along with example phrases and nugget occurrence frequencies of \dataset.  For simplicity, we use the category codes defined in the table (with brackets) to reference specific categories. 
% Formal category definitions are provided in the Appendix \S\ref{sec:appendix} \kyle{do}.

\begin{table}[t]
% \small
\footnotesize
\centering
\renewcommand{\arraystretch}{1.2}
\begin{tabular}{@{}llr@{}}
\toprule
\textbf{Category} & \textbf{Example phrase} & \begin{tabular}[c]{@{}r@{}}\textbf{\% of} \textbf{\tldrs} \\ \textsc{Auth} / \textsc{PR} \end{tabular} \\ \midrule
\begin{tabular}[c]{@{}l@{}}\textbf{[A]}rea, field \\ or topic of study\end{tabular} & \textit{\begin{tabular}[c]{@{}l@{}}reinforcement learning, \\ dependency parsing \end{tabular} } & \begin{tabular}[c]{@{}r@{}} 85.6 / 90.8 \end{tabular} \\ \hdashline[0.4pt/2pt]
\begin{tabular}[c]{@{}l@{}}\textbf{[P]}roblem or \\  motivation\end{tabular} & \textit{\begin{tabular}[c]{@{}l@{}}mode collapse,  \\ catastrophic forgetting\end{tabular}} & \begin{tabular}[c]{@{}r@{}} 29.0 / 32.9 \end{tabular} \\\hdashline[0.4pt/2pt]
\begin{tabular}[c]{@{}l@{}}Mode of \\  \textbf{[C]}ontribution\end{tabular} & \textit{\begin{tabular}[c]{@{}l@{}}method, dataset,\\ proof, theorem\end{tabular}} & \begin{tabular}[c]{@{}r@{}} 68.4 / 76.3 \end{tabular} \\\hdashline[0.4pt/2pt]
\begin{tabular}[c]{@{}l@{}}\textbf{[D]}etails or \\ description\end{tabular} & \textit{\begin{tabular}[c]{@{}l@{}}graph convolution \\ operations with dynam- \\ ically computed graphs\end{tabular}} &  \begin{tabular}[c]{@{}r@{}} 43.4  / 57.9 \end{tabular} \\ \hdashline[0.4pt/2pt]
\begin{tabular}[c]{@{}l@{}}\textbf{[R]}esults or \\ findings\end{tabular} & \textit{\begin{tabular}[c]{@{}l@{}}improved performance \\ on ImageNet, \\ simple defenses work on \\ MNIST but not CIFAR \end{tabular}} &  \begin{tabular}[c]{@{}r@{}} 29.0 / 17.1  \end{tabular} \\ \hdashline[0.4pt/2pt]
\begin{tabular}[c]{@{}l@{}}\textbf{[V]}alue or \\  significance\end{tabular} & \textit{\begin{tabular}[c]{@{}l@{}}novel, state-of-the-art, \\ simple yet effective, \\ easily applicable \end{tabular}} & \begin{tabular}[c]{@{}r@{}} 23.7 / 7.9  \end{tabular}   \\ \bottomrule
\end{tabular}
\caption{Example categories (or nuggets) of information a \tldr might contain.  Proportion of \tldrs containing each nugget estimated on 76 randomly sampled gold papers (each with its \tldr-Auth and a sampled \tldr-PR).  Percentages do not sum to one because each \tldr can contain multiple nuggets.}
\label{tab: tldr-categories-with-pcts}
\end{table}

% We observe any individual \tldr contains only a subset of the available types of information.  Counting the number of nuggets that occur per \tldr, we notice that both \auth and \peerreview typically contain \emph{two to four} nuggets, and never all six. Full counts in Appendix \S\ref{sec:num_nuggets}.

Most \tldrs contain between two to four nuggets (never all six), and will provide some indication of their subject area (\textbf{A}) and the paper's contributions (\textbf{C}).  In fact, they are the most frequently \emph{co-occurring} nuggets, appearing in 63\% of \auth and 71\% of \peerreview.  \auth tend to include results or scientific/theoretical findings (\textbf{R}) and often signal the value of their work (\textbf{V}) by describing their contributions as \emph{novel} or their results as \emph{strong} or \emph{state-of-the-art}.  In contrast, \peerreview focus more on articulating problems the paper addresses (\textbf{P}).  Interestingly, \peerreview place less emphasis on \textbf{R} and \textbf{V} in favor of further methodological details in the paper \textbf{D}.  More details about nuggets in Appendix \S\ref{sec:num_nuggets}.

\subsection{Variability in \tldrs}
\label{sec:variation}
% \includetable{gold-variation}

To explore variability in our human-written summaries, we examine differences between \tldrs written by authors (\auth) and \tldrs derived from the perspective of a peer reviewer (\peerreview). 

\begin{figure}[t]
    \centering
   \includegraphics[width=1.0\linewidth]{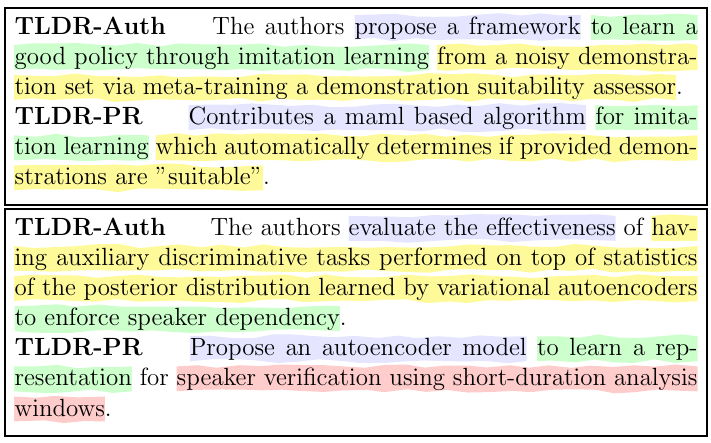}
    \caption{Two example \tldr-Auth and \tldr-PR pairs with colored spans corresponding to nuggets in Table~\ref{fig:nugget-examples} -- \hlc[red!20]{\textbf{A}}, \hlc[green!20]{\textbf{P}}, \hlc[blue!10]{\textbf{C}}, \hlc[yellow!40]{\textbf{D}}.  On \textbf{top}, we see \tldrs can have substantial lexical variation despite covering similar information content.  On \textbf{bottom}, we naturally see even more variation when the information content differs.}
    \label{fig:nugget-examples} 
\end{figure}

\paragraph{Lexical variation}
First, we note that \auth are on average 18.9 words long, while \peerreview are slightly longer on average at 22.9 words. Despite similarities in length, the 1-, 2-, and 3-gram mean Jaccard indices between \auth and \peerreview are 15.0\%, 2.5\%, and 0.7\%, respectively, indicating extremely little lexical overlap between the two sources of \tldrs. We can also observe through qualitative examples in Figure~\ref{fig:nugget-examples} how \auth and \peerreview can differ greatly, even when they contain the same information content.

%The Jaccard index is computed as follows: $J(Auth,PR) = \frac{\mid Auth \cap PR \mid}{\mid Auth \cup PR \mid}$ 
%in which $Auth$ is the set of n-grams in the author written \tldr and $PR$ is the set of n-grams in the peer review derived \tldr. 

\paragraph{Abstractiveness}
\peerreview is more abstractive with a novelty score of 20.2\% compared with \auth with a novelty score of 9.6\%, where novelty is computed as the percentage of words in the \tldr \emph{not} in the source paper. This is not unexpected because \peerreview are derived from peer review comments which themselves have already gone through one stage of abstraction.

%  KYLE -- DONT DELETE
% \begin{table}[]
%     \centering
%     \begin{tabular}{@{}lrrr@{}}
%         & 1gram & 2gram & 3gram\\
%     pr: & 20.15 &  67.22 &  89.64\\
%     author:& 9.60  &  44.54 &  69.13
%     \end{tabular}
%     \caption{Caption}\labe
%     \label{tab:my_label}
% \end{table}

\section{\ours}\label{sec:title-gen}

We introduce \ours (\fullname), a simple yet effective method for learning to generate \tldrs. Our approach addresses two main challenges: (1) the limited size of the training data and (2) the need for  domain knowledge in order to write high-quality gold \tldrs.
To address these challenges, we propose using {\em titles} of scientific papers as additional generation targets. As titles often contain key information about a paper, we hypothesize that training a model to generate titles will allow it to learn how to locate salient information in the paper that will be also useful for generating \tldrs. In addition, all papers have a title, and thus we have an abundant supply of paper-title pairs for training. 

Incorporating auxiliary \textbf{scaffold} tasks via multitask learning has been studied before for improving span-labeling and text classification \cite{Swayamdipta2018SyntacticSF, cohan-etal-2019-structural}. Similar to multitask learning, training on heterogenous data annotated with \textbf{control codes} has been shown to improve controlled generation in autoregressive language models \cite{Keskar2019CTRLAC, ElSahar2020SelfSupervisedAC, sudhakar-etal-2019-transforming, Li2020ConditionalAF}. 
% \isabel{a bunch of papers here, not sure if we need all of them}. 
In fact, it has been shown effective for generating biomedical abstracts \cite{Sybrandt2020CBAGCB}. We demonstrate that control codes can be used to effectively incorporate scaffold tasks (e.g. title generation) for denoising autoencoders like \bart \cite{Lewis2019BARTDS}. 

\begin{figure}[t!]
    \centering
    \includegraphics[width=\linewidth]{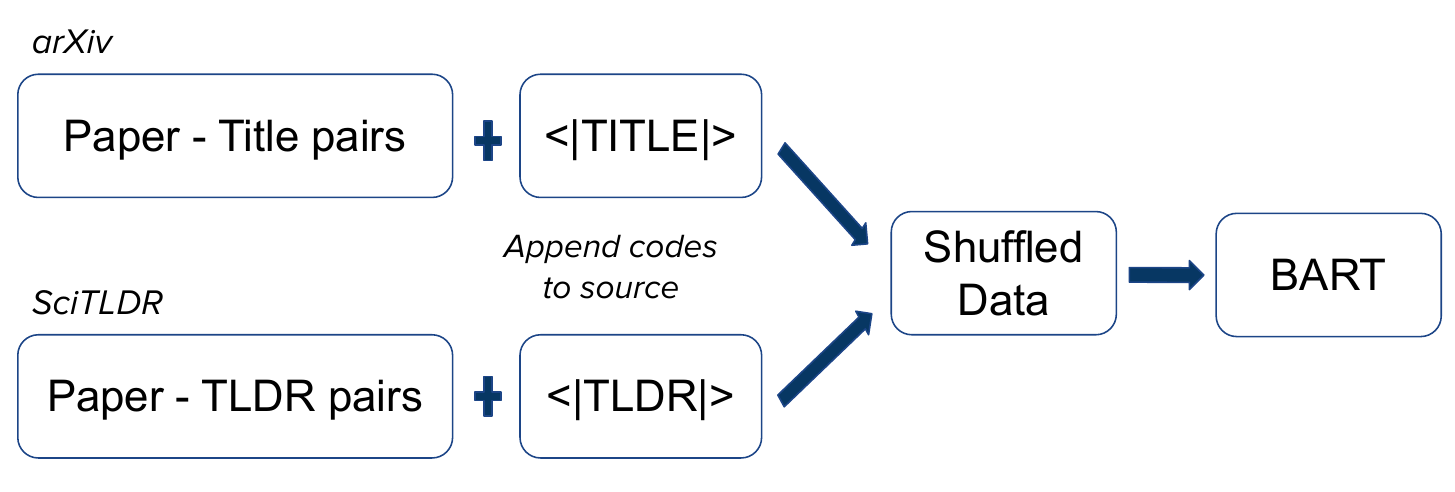}
    \caption{Training regimen for \ours.}
    \label{fig:catts}
\end{figure}

% We hypothesize that because titles are both short and contain salient information about the paper, training the model on both tasks simultaneously would be useful for the summarization model to find some of the key points of the paper. Furthermore, because all papers have titles, paper titles are easy to collect and therefore allow us to train on more data. 
In order to use title generation as a scaffold task for \tldr generation, we propose shuffling
\dataset with a title generation dataset, then appending each source with control codes $\langle\vert$\texttt{TLDR}$\vert\rangle$  and $\langle\vert$\texttt{TITLE}$\vert\rangle$, respectively. This allows the parameters of the model to learn to generate both \tldrs and titles. This process is visualized in Frigure~\ref{fig:catts}.
% , taking advantage of the close relationship between the two tasks. 
At generation time, the appropriate control code is appended to the source.  Additionally, up-sampling particular tasks can be viewed as applying task-specific weights, similar to weighting losses in multitask learning setups. 

\section{Experiments}\label{sec:experimental_setup}

% In this section we discuss our experimental setup, baselines that we use for \dataset, as well as implementation and training details.

\subsection{Baselines}
\label{subsec:baselines}

We establish baselines for \tldr generation on \dataset using state-of-the-art extractive and abstractive summarization models.

\paragraph{Extractive methods}

We consider both unsupervised and supervised extractive methods. For our unsupervised baseline, we use PACSUM \cite{Zheng2019PACSUM}, an extension of TextRank \cite{Mihalcea2004TextRankBO} that uses BERT \cite{Devlin2019BERTPO} as a sentence encoder. %For the heuristic approach, we extract a single sentence with the highest ROUGE-1 score overlap with the title. This is similar to weakly supervised data generation on most extractive summarization models \cite{Nallapati2016SummaRuNNerAR}.
For our supervised baselines, we use \bertsumext \cite{Liu2019TextSW}, which uses BERT as a sentence encoder augmented with inter-sentence Transformer layers to capture interactions,
and MatchSum \cite{matchsum}, which uses a BERT Siamese network to score whole summaries.

% These two methods are state-of-the-art on XSUM for unsupervised and supervised extractive summarization, respectively.
% \kyle{match-sum and others?}\ac{addressed}
% \kyle{can we say these are the SOTA methods on which datasets?}

\paragraph{Abstractive methods}
Since \tldrs often contain information spread across multiple sentences, we expect abstractive summarization methods to produce strong results for this task.  We focus on \bart \cite{Lewis2019BARTDS}, a Transformer-based denoising autoencoder for pretraining sequence-to-sequence models.  We use \bart-large, which achieves state-of-the-art results in summarization on XSUM. We additionally use \bart-large finetuned on XSUM, hypothesizing that the task of extreme summarization of news articles might transfer to \tldr generation on \dataset. 

\paragraph{Oracle} 

We define a sentence-level extractive oracle:  Given a paper and its multiple gold \tldrs, it selects the single sentence in the document with the highest Rouge overlap for each gold \tldr.  Then it returns the single sentence that yields the maximum Rouge across all gold \tldrs.  This sets an upper-bound on the performance of the sentence-level extractive methods under our multi-target evaluation (Section~\ref{sec:evaluation}).  Our full text oracle achieves 54.5 Rouge-1, 30.6 Rouge-2, and 45.0 Rouge-L on the test set.

\subsection{Input space}
\label{sec:input-space}

The \textbf{input space} is the context provided to the model when generating \tldrs. 

\paragraph{Abstract-only} Since the vast majority of scientific papers do not have open-access full text \cite{lo-wang-2020-s2orc}, it is  worth considering the setting in which we generate \tldrs for  papers given only their abstracts as input. The average length of an abstract is 159 words and resulting compression ratio is 7.6.

\paragraph{AIC} Previous studies have found that the most salient information in a paper for writing a summary is often found in the abstract, introduction, and conclusion (AIC) sections \cite{Sharma2019BIGPATENTAL}. An important consequence of this is the ability to substantially reduce computational costs\footnote{Especially for methods that rely on $O(n^2)$ inter-sentence comparisons or wrappers around Transformer-based methods to long contexts.} \cite{schwartz2019green} by supplying only these sections as context.  The average combined length of these contexts is 993 words and resulting compression ratio is 47.3, which is still higher than other datasets surveyed in Table \ref{tab:compare-datasets}.  

% \begin{table}[hbt!]
% \centering
% \small
% \begin{tabular}{lrrr}
% \toprule 
% \textbf{Input space} & \textbf{Rouge-1} & \textbf{Rouge-2} & \textbf{Rouge-L} \\ 
% \midrule
% Abstract-only & 47.7 & 24.7 & 38.5 \\
% \aic & 52.4 & 29.0 & 42.9 \\
% \fulltext & 54.5 & 30.6 & 45.0 \\
% \bottomrule
% \end{tabular}
% \caption{Sentence-level extractive oracle performance.}\label{tab:oracle-results}
% \end{table}

Comparing oracle results in Table~\ref{tab:results}, we see that increasing the input space from abstract-only to \aic improves Rouge-1 by +4.7.  Yet, this is only 2.1 Rouge-1 lower than the full text oracle performance, despite requiring five times more text.

% To keep our experimental setup simple, we restrict to using abstracts (\ao) and the concatenation of abstract, intro, and conclusion (\aic) as candidate input spaces for \ours and the baselines.

% {\color{red} Maybe footnote somewhere, but for abstractive methods, could've also defined input space as top extractions from an extractive method.  We didn't do this because results were super far behind the oracle, and simply providing the full \aic seems reasonable given the Oracle analysis.}

%   KYLE -- CANT REALLY SAY THIS
% Most pretrained Transformer models have an input length constraint and require significant customization when applied to long contexts, like scientific papers \kyle{cite}. \kyle{BART supports how much?  BERTSumExt supports how much?} 

%  KYLE -- WHY ABSTRACTS CAN BE USEFUL?

\setlength\dashlinedash{0.2pt}
\setlength\dashlinegap{1.5pt}
\setlength\arrayrulewidth{0.3pt}

\begin{table*}[t]
\centering
\small
\renewcommand{\arraystretch}{1.1}
\begin{tabular}{@{}lrrrrrr@{}}
\toprule
   & \multicolumn{3}{c}{\textbf{Abstract-only}} & \multicolumn{3}{c}{\textbf{\aic}} \\ 
 \cmidrule(lr){2-4} \cmidrule(lr){5-7}
\textbf{Method}   & R1 & R2 & RL & R1 & R2 & RL \\ \midrule
\emph{Oracle}  & 47.7 & 24.7  & 38.5 & 52.4  & 29.0   & 42.9  \\
PACSUM \cite{Zheng2019PACSUM}    & 19.3                     & 4.0                  & 15.1                  & 28.7                     & 9.8                  & 21.9                  \\
\bertsumext \cite{Liu2019TextSW}  & 38.5                     & 16.6                  & 30.5                  & 36.2                     & 14.7                  & 28.5                  \\
MatchSum \cite{matchsum}  & 42.7                & 20.0                  & 34.0                  & 38.6                     & 16.4                  & 30.1                  \\
\hdashline\noalign{\vskip 0.5ex} 
\bart \cite{Lewis2019BARTDS}      & 43.3                     & 20.8                  & 35.0                  & 42.9                     & 20.8                  & 35.1                  \\
\bartxsum \cite{Lewis2019BARTDS} & 42.5                     & 21.1                  & 34.9                  & 43.7                     & 21.4                  & 36.0                  \\
\hdashline\noalign{\vskip 0.5ex} 
\ours (Ours)                     & \bf{43.8}         & 20.9                  & 35.5           & $^\dag$\bf{44.9}         & $^\dag$\bf{22.6}             & $^\dag$\bf{37.3}             \\
\oursxsum (Ours)                & $^\dag$44.3              & \bf{21.3}             & \bf{35.9}             & 44.6                     & 21.7                  & 36.5                  \\ \bottomrule
\end{tabular}
\caption{Test set max Rouge scores of extractive and abstractive baselines and \ours. We use $^\dag$ to indicate \ours variants that significantly ($p$$<$$0.05$) outperform their corresponding \bart baseline.}\label{tab:results}
\end{table*}

\subsection{Training and implementation details}

All experiments use Titan V or V100 GPUs. 
% Below, we describe the implementaion details for the baselines and \ours. 
We experiment on abstract-only and \aic input spaces.
Best hyperparameters for the models are selected based on dev set Rouge-1.  Supervised models like \bertsumext and \bart are trained on \dataset and the 
% trained for up to 5 epochs and 
best model checkpoint chosen using dev set loss. See Appendix\S\ref{sec:additional_training_details} for additional parameter tuning details of all models.

\paragraph{Extractive Methods}
    For PACSUM, \bertsumext and MatchSum we use original code released by the authors. The first two use BERT-base and the last one uses RoBERTa-base \cite{liu2019roberta}.  
    For MatchSum in AIC input space, following the authors, we use \bertsumext to first extract 7 highly scoring sentences as the input to MatchSum.\footnote{In abstract-only setting, MatchSum takes the full context.}
    Sentence segmentation is performed using ScispaCy \cite{Neumann2019ScispaCyFA}, and models select a single sentence as their predictions. We use the default hyperparameters for PACSUM.

\paragraph{Abstractive Methods}
    We experiment with \bart-large and \bart-large finetuned on XSUM, using the Fairseq \cite{ott2019fairseq} implementation and the released XSUM weights.
    % \footnote{\url{https://github.com/pytorch/fairseq}} 
    We apply the \ours training method to these two models, using an additional 20K paper-title pairs from arXiv for title generation.\footnote{Includes all papers on arXiv with at least one of the following tags \textsc{cs.CL}, \textsc{cs.CV}, \textsc{cs.LG}, \textsc{cs.AI}, \textsc{cs.NE}, and \textsc{stat.ML} \emph{and} have identified introduction and conclusion sections by S2ORC \cite{lo-wang-2020-s2orc}.} We up-sample \tldr instances to match the size of the title scaffold data.\footnote{While this up-sampling may indicate that \ours is training on more \tldrs than \bart, we allow \bart training up to 20 epochs and it quickly overfits within a few epochs.}
    % \kyle{How many epochs did \ours actually need?}}  
    For simplicity, we refer to these as \bart, \bartxsum, \ours and \oursxsum, respectively. For all models, we use a learning rate of 3e-5, update frequency of 1, and max tokens per batch of 1024\footnote{Fairseq reports an ``average batch size'' of 36, which is a consequence of adaptive batching of examples based on the update frequency and max tokens per batch.} chosen through manual tuning.  We tune decoder for all models via grid search over five length penalties between 0.2 and 1.0 and 7 beam sizes 2 to 8.
    
    % \footnote{Using the provided hyperparameter search script, we did not observe improvement over the defaults.} 

    % Due to model input constraints, we experiment on abstract-only and \aic input spaces, but provide an additional analysis in \S\ref{sec:analysis} exploring the potential benefits of full text.

\subsection{Evaluation}
\label{sec:evaluation}

\paragraph{Automated evaluation}

% We use the {\color{red} better justification?} current standard evaluation metric in summarization, the Rouge \cite{lin-2004-rouge} framework. 

Following recent work on extreme summarization \cite{narayan-cnn-xsum,Lewis2019BARTDS}, we use Rouge-1, Rouge-2, and Rouge-L \cite{lin-2004-rouge} as our automated metrics. As discussed in Section~\ref{sec:dataset}, we have multiple target summaries available per paper.  To exploit this during evaluation, we calculate the Rouge score of the system-generated \tldr with respect to each of the gold \tldrs for the corresponding paper (including its \tldr-Auth and all of its \tldrs-PR) individually.  We take the \textbf{maximum} Rouge score over these gold \tldrs as the final Rouge score for that paper. An alternative approach to aggregating scores would be to take the mean, but due to the variability in \tldrs shown in Section~\ref{sec:variation}, we argue the maximum operation is more appropriate -- That is, matching \emph{any} of the gold \tldrs is rewarded.\footnote{For completeness we provide mean Rouge scores in Appendix Table~\ref{tab:mean-r-results} to supplement our main max Rouge results in Table~\ref{tab:results}.}

\paragraph{Human evaluation}
While our multi-target setting allows us to mitigate some of the limitations of Rouge \cite{conroy2011nouveau,Cohan2016RevisitingSE}, we acknowledge that relying only on automated metrics is insufficient for evaluating the quality of the models.  In addition to automated metrics, we also have human experts in computer science assess system-generated \tldrs under two criteria -- informativeness and correctness. 

For \textbf{informativeness}, we perform the nugget-based analysis for information content over system-generated \tldrs for the same 76 gold papers from Section~\ref{sec:information_content}.  We use the presence (or lack) of different nuggets in predicted and gold \tldrs to quantify differences in information content.  Specifically, we score each gold and system-generated \tldr by \emph{the number of unique nuggets divided by the number of tokens}. This length normalization handles cases where systems returning the source document are trivially more informative.  For each paper, we rank the predicted and gold \tldrs.  Then, we compute overall metrics for each gold or system variant by aggregating their ranks across papers using mean reciprocal rank (MRR).

Evaluating \textbf{correctness} requires careful reading and understanding the source paper.  To minimize this burden and have reliable evaluation, we ask the original authors of papers to assess the correctness of our system-generated \tldrs.
% assuming authors are able to identify factual incorrectness with minimal effort when it pertains to statements about their own work.  \ac{commented this for space}
We manually email (first or second) authors of arXiv papers and ask them to score each system-generated \tldr with \emph{1 - false or misleading}, \emph{2 - partially accurate} or \emph{3 - mostly correct}, regardless of comprehensiveness.  We compare the mean correctness (across papers) for each system variant.  We received responses from 29 unique authors with annotations covering 64 arXiv papers.

\section{Results}\label{sec:results}

\subsection{Quantitative results}

We present our main results in Table~\ref{tab:results}.  

\paragraph{Extractive results}

We establish baseline results for extractive methods on our new dataset \dataset. 
%  KYLE -- NO MATCHSUM RESULTS
We observe that MatchSum has the highest extractive performance, followed by \bertsumext. 
We observe that increasing input space from abstract-only to \aic greatly improves PACSUM\footnote{PACSUM using the full text yields a Rouge-1 of 12.7, significantly worse than abstract-only.} performance but decreases performance of both \bertsumext and MatchSum. We suspect that increasing the input space makes it more difficult for these models to learn optimal parameters including new position embeddings in low-resource training. Compared to the extractive oracle scores, we see there is plenty of room for improvement.

\begin{table}[t]
\centering
\footnotesize
\begin{tabular}{@{}lrrrr@{}}
\toprule
& MRR   &  \begin{tabular}[c]{@{}l@{}}Avg. \# \\ nuggets \end{tabular}  & \begin{tabular}[c]{@{}l@{}}Avg. \#  \\ words \end{tabular} \\ \midrule
\tldr-Auth (Gold)          & 0.53 & 2.5 & 20.5  &         \\   % 127 total
\tldr-PR (Gold)            & 0.60 & 2.4 & 18.7  &         \\  % 121 total
\bartxsum                  & 0.42 & 2.2  & 19.4  &         \\  % 111
\oursxsum                  & 0.54 & 2.6 & 20.8  &         \\  % 132
\bottomrule
\end{tabular}
\caption{Human evaluation on informativeness of gold and system-generated \tldrs. Higher MRR corresponds to variants that, on average, rank higher than others by length-normalized number of nuggets.}\label{tab:human-eval}
\end{table}

% % % % % % % % % % % 

\begin{table}[t]
\footnotesize
\begin{tabular}{@{}lrrrrrr@{}}
\toprule
   & \multicolumn{2}{c}{\textbf{Abstract-only}} & \multicolumn{2}{c}{\textbf{\aic}} \\ 
 \cmidrule(lr){2-3} \cmidrule(lr){4-5}
\textbf{Method}   & 
\begin{tabular}{@{}c@{}}
  \% novel \\ words
\end{tabular}
& 
\begin{tabular}{@{}c@{}}
  Avg. \# \\ words
\end{tabular}
&
\begin{tabular}{@{}c@{}}
  \% novel \\ words
\end{tabular}
&
\begin{tabular}{@{}c@{}}
  Avg. \# \\ words
\end{tabular}
\\ \midrule
\bart & 2.9\% & 20.9 & 1.3\% & 20.4  \\
\bartxsum & 3.7\% & 18.4 & 1.1\% & 18.9  \\
\hdashline\noalign{\vskip 0.5ex} 
\ours & 5.5\% & 19.1 & 5.3\% & 18.4  \\
\oursxsum & 5.8\% & 19.7 & 4.5\% & 19.7  \\
\bottomrule
\end{tabular}
\caption{Lexical features of system-generated \tldrs. }
\label{tab:analysis}
\end{table}
% % % % % % % % % % % 

% % % % % % % % % % % 
%  KYLE -- Moving this table for Analysis around to get it right position
\begin{table}[t]
\centering
\footnotesize
\begin{tabular}{@{}lrrrrrr@{}}
\toprule
 \textbf{Method}     & \textbf{R1} & $\Delta$  & \textbf{R2} & $\Delta$  & \textbf{RL} & $\Delta$  \\ \midrule
\bart  & 44.9 & +1.6 & 22.6 & +1.8 & 37.1 & +2.1 \\
\bartxsum  & 44.8 & +1.1 &21.8 & +0.4 & 36.4 & +0.4 \\
\ours & 44.9 & +0.0 & 21.9 & -0.7 & 36.6 & -0.7 \\ 
\oursxsum & 45.7 & +1.1 & 23.0 & +1.7 & 37.1 & +1.2 \\ 
\bottomrule
\end{tabular}
\caption{Oracle input space experiments.  $\Delta$ are differences between oracle result and model's best performance (across abstract-only and \aic) from Table \ref{tab:results}.}
\label{tab:oracle_input_source}
\end{table}
% % % % % % % % % % % 

\paragraph{Abstractive results}  

Abstractive methods are not limited to choosing exact sentences. 
% We observe that our BART baselines slightly outperform the best extractive baseline (MatchSum). \kyle{no longer have matchsum} 
For a given abstractive baseline \bart or \bartxsum, our \ours learning strategy results in improvements in both abstract-only and AIC settings. 
Comparing \ours variants with their corresponding \bart baselines, we observe that in the abstract-only setting, \ours and \oursxsum achieve +0.5 and +1.8 Rouge-1, respectively. In the \aic setting, \ours and \oursxsum achieve +2.0 and +0.9 Rouge-1, respectively.  We use the two-sided paired t-test against a null hypothesis of no difference to assess these differences. To address the issue of multiple hypothesis testing over Rouge scores, we perform a Holm-Bonferroni \cite{holm79}\footnote{Using the \textsc{p.adjust} library in \textsc{R} \cite{rlanguage}} correction for determining significant $p$-values in Table~\ref{tab:results}.

\subsection{Human evaluation}
\label{sec:human_eval}

We perform our human evaluation on \bartxsum and \oursxsum using the \aic input space on 51 sampled papers. In this setting, we have both chosen the strongest baseline and controlled for XSUM pretraining. From Table~\ref{tab:human-eval}, we see that \oursxsum is more informative than \bartxsum and is comparable to gold \auth, though still less informative than \peerreview. 

In addition to informativeness, we also evaluate content accuracy of generated tldrs as explained in Section~\ref{sec:evaluation}. 
We report no difference in correctness between \bartxsum and \oursxsum.  We observe 42 ties, 10 cases where \bartxsum is more correct, and 12 cases where \oursxsum is more correct.  Both models average a rating of 2.5 (scoring between partially accurate and mostly correct).

\subsection{Analysis}
\label{sec:analysis}

\paragraph{How abstractive are the generations?}

From Table~\ref{tab:analysis}, we observe: (1)  \bart variants are less abstractive than \ours variants. (2) Initial training on XSUM might influence models to be slightly less abstractive. (3) \bart variants are more abstractive in the abstract-only setting than the longer \aic settings, while \ours seems to have the same level of abstractiveness regardless of input space.

\paragraph{How long are the generations?}
From Table~\ref{tab:analysis}, we see the systems all generate \tldrs of similar length to the average length reported in Table \ref{tab:compare-datasets}.

\paragraph{How important is using the full text?}
To analyze whether one can improve abstractive model performance by improving the input space selection (compared to just using \aic), we define an \emph{oracle input space}. That is, for each \tldr, we select sentences from the full text that maximize Rouge-1 with the gold \tldrs-Auth\footnote{Only \tldrs-Auth is exists for all papers. \tldrs-PR are only in dev and test.} and select the top sentences to match the length of \aic.  Repeating the experiments in Section~\ref{sec:experimental_setup} with this input source, we observe some performance improvement across models (Table~\ref{tab:oracle_input_source}).

\paragraph{Qualitative example}
Table \ref{tab:qualitative-example} contains system generations on the same paper (alongside the gold \tldrs). Curiously, despite both achieving the same Rouge-1, the generated \tldrs are quite different.  \bartxsum focuses on the methodological contribution while \oursxsum focuses on a scientific finding.  The ``two hidden layer'' detail by \bartxsum is from the paper introduction and the ``defining the appropriate sampling distributions'' from \oursxsum is from the conclusion.\footnote{See original paper: \\ \href{https://openreview.net/pdf?id=SkGT6sRcFX}{https://openreview.net/pdf?id=SkGT6sRcFX}}

\begin{table}
\centering
\small
\setlength{\tabcolsep}{2pt}
\begin{tabular}{@{}p{\linewidth}@{}}
\toprule
 \textbf{\tldr-Auth} \hspace{1em} We propose a method for the \hlc[orange!40]{construction} of arbitrarily deep \hlc[green!20]{infinite-width networks}, based on which we derive a novel \hlc[blue!20]{weight initialisation} scheme for finite-width networks and demonstrate its competitive performance. \\ 
 \hdashline\noalign{\vskip 0.5ex} 
 \textbf{\tldr-PR} \hspace{1em}  Proposes a \hlc[blue!20]{weight initialization} approach to enable infinitely deep and \hlc[green!20]{infinite-width networks} with experimental results on small datasets. \\
 \midrule
\textbf{\bartxsum} \hspace{1em}  We propose a principled approach to \hlc[blue!20]{weight initialisation} that allows the \hlc[orange!40]{construction} of \hlc[green!20]{infinite-width networks} with more than two hidden layers. \\
\hdashline\noalign{\vskip 0.5ex} 
\textbf{\oursxsum} \hspace{1em}   We study the \hlc[blue!20]{initialisation} requirements of \hlc[green!20]{infinite-width networks} and show that the main challenge for \hlc[orange!40]{constructing} them is defining the appropriate sampling distributions for the \hlc[blue!20]{weights}. \\
\bottomrule
\end{tabular}
\caption{Examples of system generations.  \bartxsum and \oursxsum both achieve Rouge-1 of 40.7 on this paper.  Colored spans indicate text overlap.
% The first example illustrates an observed phenomenon with \bart and \bartxsum which generates background/motivational sentences in approx. 5-10\% of cases. 
% \kyle{Is it better to have another example, or to have \bart and \ours (no xsum)?} 
} \label{tab:qualitative-example}
\end{table}

\section{Related work}
\label{sec:related_work}

\paragraph{Transformers for summarization} Transformer-based models have achieved strong results in extractive and abstractive summarization. 
PACSUM \cite{Zheng2019PACSUM} combines BERT sentence representation with unsupervised text ranking; MatchSum \cite{matchsum} uses a Siamese BERT model to score the entire summary instead of a single extraction; and \citet{Liu2019TextSW} show that BERT is effective for both extractive and abstractive summarization. \citet{Zhang2019PEGASUSPW,Bi2020PALMPA} introduce new pretraining objectives that improve generation.  Sequence-to-sequence models \cite{T5-Raffel2019ExploringTL,Lewis2019BARTDS,Bao2020UniLMv2PL} have state-of-the-art performance on XSUM \cite{narayan-cnn-xsum}, a dataset for extreme summarization dataset of news articles.  \dataset is a new form of extreme summarization focused on scientific papers.

\paragraph{Scientific document summarization}
% Information overload has resulted in considerable interest in summarization datasets and methods for scientific documents in recent years. 

% SciSummNet with human-written reference summaries \cite{Yasunaga2019ScisummNetAL}; the CL-SciSumm \cite{jaidka2018insights,scisumm-chandrasekan} datasets which incorporate citation contexts, and 

Most work in summarization of scientific papers have focused on longer summaries (i.e. 150-200 words). Existing datasets include CSPubSum for extractive summarization \cite{Collins2017ASA}, ArXiv and PubMed for abstract generation \cite{Cohan2018ADA}, and SciSummNet \cite{Yasunaga2019ScisummNetAL} and CL-SciSumm \cite{jaidka2018insights,scisumm-chandrasekan} datasets, which incorporate citation contexts into human-written summaries. TalkSumm \cite{Lev2019TalkSummAD} uses recordings of conference talks to create a distantly-supervised training set for the CL-SciSumm task. 
% \kyle{please check; i rephrased the scisumm ones}\ac{looks good}

Modeling approaches in scientific document summarization include models that exploit citation contexts \cite{Qazvinian2013GeneratingES,Cohan2015ScientificAS,cohan2017scientific, Zerva2020CitedTS}, 
% generating citation contexts, given two papers \cite{Luu2020CitationTG}, 
automated survey generation \cite{Mohammad2009UsingCT,Jha2015SurveyorAS,fabbri-etal-2018-tutorialbank,wang2018citationas}, and other techniques focusing on exploiting the unique properties of scientific documents such as long length and structure \cite{Conroy2017SectionMM,Nikolov2018DatadrivenSO,Cohan2018ADA, Xiao2019ExtractiveSO}.  Yet, such methods have not been studied in the setting of extreme summarization (i.e. short target summaries, high compression, high abstraction), and \dataset is the first dataset to facilitate such research.

\section{Conclusion}

% \ac{can we make this a bit punchier? Like we should probably add some implications of our work. e.g., why people should care, why tldrs are intersesting.}

We introduce \tldr generation for scientific papers, and release \dataset, a multi-target dataset of \tldr-paper pairs.  We also present \ours, a simple yet effective learning strategy for improving \tldr generation that exploits auxiliary training signal from paper titles.  We show that our approach improves over strong modeling
% abstractive 
% \ac{and extractive} 
% \kyle{we never used CATTS with extractive}
baselines.  

Existing methods for scientific document summarization often make use of properties unique to those papers, like sections, citation contexts or scientific discourse roles.  Future work can examine how best to incorporate these properties to improve \tldr generation models. Additionally, while our experiments are limited to abstract-only and AIC input spaces, we provide the full text of the source papers to support research into using longer input contexts.  Furthermore, the multiple target summaries in \dataset reflect diverse perspectives and can be used to support summarization research into training and evaluation techniques previously unavailable with existing datasets.  Finally, the idea of a \tldr can differ between academic disciplines, and we leave such exploration open for future work.

\section*{Acknowledgments}
We thank the Semantic Scholar Research team and John Bohannon and Oleg Vasilyev from Primer for helpful feedback and discussions.
This work was supported in part by NSF Convergence Accelerator
award 1936940, NSF RAPID award 2040196,
ONR grant N00014-18-1-2193, and the University
of Washington WRF/Cable Professorship.

\bibliography{emnlp2020}
\bibliographystyle{acl_natbib}

\appendix
\clearpage

% \appendix
% \section*{Appendix}
\label{sec:appendix}

% \section{Category definitions}
% \label{sec:category_definitions}

% {\color{red} todo}

\section{How many nuggets in \tldrs?}
\label{sec:num_nuggets}

\begin{table}[hbt!]
    \centering
    \footnotesize
    \begin{tabular}{l|llll}
    \toprule
    \textbf{\# categories} & \textbf{0} & \textbf{1} & \textbf{2} & \textbf{3} \\
    \hdashline[0.5pt/0.5pt]\noalign{\vskip 0.5ex}
    \tldr-Auth & 2.6\% & 10.5\% &  26.3\% & 34.2\%  \\
    \tldr-PR & 0.0\% & 9.2\% & 30.3\% & 31.6\% \\
    \bottomrule
    \textbf{\# categories} & \textbf{4} & \textbf{5} & \textbf{6}  \\
    \hdashline[0.5pt/0.5pt]\noalign{\vskip 0.5ex}
    \tldr-Auth  & 18.4\% & 7.9\% & 0.0\% \\
    \tldr-PR &  26.3\% & 2.6\% & 0.0\% \\
    \bottomrule
    \end{tabular}
    \caption{Number of categories represented in a \tldr}
    \label{tab:num_nuggets}
\end{table}

% KYLE -- add somewhere later
% \section{Processing of \tldrs}

% Some \tldrs are written in first person, which are converted using regular expressions to third person.  This includes transformations like ``we'' to ``the authors'' and ``our'' to ``their.''

\section{Breakdown of venues in \dataset?}
\begin{table}[hbt!]
    \centering
    \footnotesize
    \begin{tabular}{ll}
    \toprule
    \textbf{Venue} & \textbf{Proportion} \\
    \hdashline[0.5pt/0.5pt]\noalign{\vskip 0.5ex}
    ICLR &                   85.2\% \\
NeurIPS/NIPS      &               5.8\% \\
OpenReview   &            2.1\% \\
ICML         &            2.0\% \\
ICAPS   &      1.8\% \\
other & 3.1\% \\
% approximateinference    & 0.014425 \\
% AKBC            &         0.007678 \\
% graphicsinterface   &     0.006980 \\
% roboticsfoundation   &    0.001861 \\
    \bottomrule
    \end{tabular}
    \caption{Breakdown of venues represented by papers in \dataset}
    \label{tab:venue_breakdown}
\end{table}

\section{Background knowledge for \tldrs}
\label{sec:background_knowledge}

What a paper's \tldr looks like or what information it should include is subjective and follows (community-specific) commonsense rather than any formally-defined procedure. Since \tldrs are inherently ultra-short, they are not necessarily self-contained statements, and understanding them requires background expertise within their respective scientific domain. Therefore, when designing \dataset, we assume readers have sufficient background knowledge to follow a general research topic in a given domain. This eliminates the need for \tldrs to include explanations or clarifications of common domain-specific terms  (e.g., ``bounds,'' ``LSTM,'' or ``teacher''). 

% \Bug{I thought the discussion (below) of what makes a good tldr helpul was interesting, but I might emphasize the role of background info: an abstract is more self contained and helpful to anyone. The reason that the scibert tldr is ‘better’ is because ‘everyone’ knows what bert is and what problem it solves so no need to repeat that.} 

\section{Additional model training details}
\label{sec:additional_training_details}

\paragraph{PACSUM}
    The default hyperparameters are beta and lambda1 set to 0. We did some initial tuning of the hyperparameters using the provided tuning code, which performs a search over 10 beta values and 10 lambda1 values.  This did not result in a significant difference in performance. PACSUM had a total runtime of 12 minutes on abstracts and 6.5 hours on \aic. 
    We used the released code by authors.\footnote{\url{https://github.com/mswellhao/PacSum}} 
    
\paragraph{\bertsumext}
    We trained with a batch size of 1 sentence per batch and for 5,000 total steps for a total training time of 30 min. We use a learning rate of 2e-3 and a dropout rate of 0.1, which are the reported parameters used for XSUM. BERTSumExt also requires a max token length for initializing position embeddings.  For the abstract-only setting, we use the default number of max tokens 512, which fits the full length of all of abstracts in \dataset.  For \aic, we first attempted 3 different truncation lengths -- 1024 (double the max tokens for abstracts), 1500 (90th percentile length), and 1800 (95th percentile length) tokens.  We found that truncation at 1500 performs best on \aic. We used the released code by authors.\footnote{\url{https://github.com/nlpyang/PreSumm}}

\paragraph{MatchSum}
    We trained MatchSum with a batch size of 32, learning rate of 2e-5 with a linear warmup and decay scheduler, and trained the model for 15 epochs. We chose the best checkpoint based on linear combination of Rouge-1, Rouge-2 and Rouge-L. We manually tuned hyperparameters -- For learning rate, we tried 2e-5 and 3e-5  and for number of epochs, we tried 5, 15, and 20. For \aic, as MatchSum requires few salient sentences as input for candidate generation, we used \bertsumext to score sentences and chose the top 7 ones as input to MatchSum.  This is according to instructions by authors\footnote{\url{https://github.com/maszhongming/MatchSum}}. Instead of training the model from scratch we used the authors released checkpoint based on the CNN/DM dataset. This resulted in about 1 Rouge-1 point improvement.
    % In addition, we found the RoBERTa checkpoint by authors to work slightly better the BERT.

\paragraph{\bart}
    For \bart and \bartxsum finetuning experiments, we train all the models for 500 steps with 20\% warm-up for an approximate training time of 45 minutes. This is equivalent to 5 epochs, though we initially allowed \bart to train for up to 20 epochs and found that the model quickly overfits to the training set (as evidenced by poor performance on the dev set).  

    Through manual tuning, we achieved the best results by reducing the training time. Also in manual tuning, we first ran the experiments on four learning rates, 2e-5, 3e-5, 4e-5, and 5e-5 and controlled for all other hyperparameters. We then tested three different seeds, again controlling for all other parameters. Finally, we tested two batch sizes, 2048 tokens per batch and 1024 tokens per batch. 

\paragraph{\ours}
    In the abstract-only setting, we train \ours  for 11,000 total steps for a total training time of 2.5 hours. For \aic, we train \ours for 45,000 total steps for a total training time of 10 hours. This also equivalent to 5 epochs of training. We do not perform tuning on the training hyperparameters for \ours, instead opting to use the same parameters as the baseline BART models. 
    
    % We then tune the decoder for all the BART models by performing a grid search over 5 length penalties (0.2, 0.4, 0.6, 0.8, 1.0) and 7 beam sizes (2 through 8).

    % \arman{emnlp asks for reporting range of hyperparameters that was tried and total number of gpu hours, not needed now, but we need to add later} \isabel{range of hyperparameters is there, need to add gpu hours}

\section{Mean ROUGE test results}

\setlength\dashlinedash{0.2pt}
\setlength\dashlinegap{1.5pt}
\setlength\arrayrulewidth{0.3pt}

\begin{table}[hbt!]
\centering
\setlength{\tabcolsep}{4pt}
\small
\begin{tabular}{@{}lrrrrrr@{}}
\toprule
   & \multicolumn{3}{c}{\textbf{Abstract-only}} & \multicolumn{3}{c}{\textbf{\aic}} \\ 
 \cmidrule(lr){2-4} \cmidrule(lr){5-7}
\textbf{Method}   & R1 & R2 & RL & R1 & R2 & RL \\ \midrule
% \emph{Oracle}  & 47.7 & 24.7  & 38.5 & 52.4  & 29.0   & 42.9  \\
% PACSUM \cite{Zheng2019PACSUM}    & 19.3                     & 4.0                  & 15.1                  & 28.7                     & 9.8                  & 21.9                  \\
% \bertsumext \cite{Liu2019TextSW}  & 38.5                     & 16.6                  & 30.5                  & 36.2                     & 14.7                  & 28.5                  \\
% MatchSum \cite{matchsum}  & 42.7                & 20.0                  & 34.0                  & 38.6                     & 16.4                  & 30.1                  \\
\bart 
% \cite{Lewis2019BARTDS}      
& 31.1 & 10.7 & 24.4 & 30.7 & 10.6 & 24.4 \\
\bartxsum 
% \cite{Lewis2019BARTDS}
& 30.1 & 10.7 & 24.1 & 31.0 & 10.9 & 24.7 \\ \hdashline
\ours 
% (Ours)   
& 31.5 &  11.0 & 24.9 & $^\dagger$31.9 & $^\dagger$\bf{11.8} & $^\dagger$\bf{25.6} \\
\oursxsum 
% (Ours) 
& $^\dagger$\bf{31.7} & \bf{11.1} & $^\dagger$\bf{25.0} & $^\dagger$\bf{32.1} & $^\dagger$11.6 & $^\dagger$25.4 \\\bottomrule
\end{tabular}
\caption{Test set results using mean Rouge scores instead of max for abstractive methods.  We use $\dagger$ to indicate \ours variants that significantly ($p < 0.05$) outperform their corresponding \bart baseline.}\label{tab:mean-r-results}
\end{table}

% \subsection{\ours on XSUM}
% {\color{red} INSERT SOMETHING HERE}

\section{\peerreview annotation instructions}
\label{sec:annotation_instructions}
Below are the instructions provided to annotators rewriting peer-review comments.

\paragraph{Task:} We want to collect a dataset of short summaries of CS papers,  but it’s hard to get people to read and write summaries about entire papers. Instead, we collected a dataset of peer reviewer comments, in which many CS researchers have read and written reviews of papers. Often, a reviewer’s comments will also include a summary of the paper they’ve read.  Our task is given the title and first 128 words of a reviewer comment about a paper, re-write the summary (if it exists) into a single sentence or an incomplete phrase. Summaries must be no more than one sentence. Most summaries are between 15 and 25 words. The average rewritten summary is 20 words long.

\paragraph{What might be included in your re-write?}
\begin{enumerate}
    \item What subfield is their work in?
    \item What problem are they trying to solve?
    \item What did the paper do?
    \item Why should you care/how is it novel?
\end{enumerate}

% \paragraph{Example:} Below is a reviewer comment about a paper:
% \begin{quote}
%     Good work
    
%     The authors propose a method for learning node representations which, like previous work (e.g. node2vec, DeepWalk), is based on the skip-gram model. However, unlike previous work, they use the concept of shared neighborhood to define context rather than applying random walks on the graph. The paper is well-written and it is quite easy to follow along with the discussion. This work is most similar, in my opinion, to node2vec. In particular, when node2vec has its restart probability set pretty high, the random walks tend to stay within the local neighborhood (near the starting node). The main difference is in the sentence construction strategy. Whereas node2vec may sample walks that have context windows containing the same node, the proposed method does not as it uses a random permutation of...
% \end{quote}

% Notice the comment is saying that the paper is about a new method for “learning node representations”, which is a subfield, and describes the “concept of shared neighborhood” that wasn’t used in previous work, which describes why it’s interesting/novel.  

% From here, we can re-write the reviewer comment to form a short summary:
% \begin{quote}
%     A method for learning node representations using the concept of shared neighborhood to define context
% \end{quote}

\paragraph{What to exclude when re-writing a comment:}
Not everything in the reviewer comment belongs in the summary.  We purposefully leave out:
\begin{itemize}
    \item Reviewer decisions/opinions (accept, reject, suggestions, etc.)
        \begin{itemize}
            \item ``The paper is well-written and it is quite easy to follow along with the discussion."
        \end{itemize}
    \item Background information/ previous work
        \begin{itemize}
            \item ``The authors propose a method for learning node representations which, like previous work (e.g. node2vec, DeepWalk), is based on the skip-gram model."
            \item ``In particular, when node2vec has its restart probability set pretty high, the random walks tend to stay within the local neighborhood (near the starting node)."
        \end{itemize}
    \item Excessive details about methodology 
        \begin{itemize}
            \item ``Whereas node2vec may sample walks that have context windows containing the same node, the proposed method does not as it uses a random permutation of..."
        \end{itemize}
\end{itemize}

% \paragraph{More examples of good summaries:} \isabel{we provided this many examples to the annotator but i'm not sure we need that many for the appendix}
% \begin{itemize}
%     \item The authors design an adversarial training method to Bayesian neural networks, showing a much stronger defense to white-box adversarial attacks
%     \item This paper proposes a special weakly-supervised multi-label learning problem along with a newly tailored algorithm that learns the underlying classifier by learning to assign pseudo-labels.
%     \item A new non-adversarial feature matching-based approach to train generative models that achieves state-of-the-art results.
%     \item The authors introduce Variational Intrinsic Successor FeatuRes (VISR), a novel algorithm which learns controllable features that can be leveraged to provide fast task inference through the successor features framework.
% \end{itemize}

\paragraph{Enter ``None" for the summary for the following conditions:}
\begin{itemize}
    \item The comment is entirely the reviewer’s opinions about the paper
    \item The reviewer’s summary carries heavy sentiment about the paper
        \begin{itemize}
            \item ``This paper presents a method that is not novel or interesting"
            \item This applies when the sentiment is so heavy that you are unable to write a summary.
        \end{itemize}
    \item If the comment is about a paper that is out of your domain of expertise.
\end{itemize}

\end{document}